\documentclass[letterpaper, 10 pt, conference]{ieeeconf}  

\IEEEoverridecommandlockouts                              
\usepackage{graphicx}
\graphicspath{ {./figures/} }

\title{\LARGE \bf
Combining Sequential and Aggregated Data for Churn Prediction in Casual Freemium Games
}

\author{Jeppe Theiss Kristensen and Paolo Burelli
\thanks{The authors are with the IT University of Copenhagen and with Tactile Games ApS. (e-mail: jeppek@tactile.dk; pabu@itu.dk).
A special thanks goes to Thomas Bjarke Heiberg-Iürgensen and Rune Viuff Petersen for their MSc thesis work.}%
}

\begin{document}


\maketitle
\thispagestyle{empty}
\pagestyle{empty}

\begin{abstract}

In freemium games, the revenue from a player comes from the in-app purchases made and the advertisement to which that player is exposed. The longer a player is playing the game, the higher will be the chances that he or she will generate a revenue within the game.
Within this scenario, it is extremely important to be able to detect promptly when a player is about to quit playing (churn) in order to react and attempt to retain the player within the game, thus prolonging his or her game lifetime.
In this article we investigate how to improve the current state-of-the-art in churn prediction by combining sequential and aggregate data using different neural network architectures.
The results of the comparative analysis show that the combination of the two data types grants an improvement in the prediction accuracy over predictors based on either purely sequential or purely aggregated data.

\end{abstract}

\section{INTRODUCTION}

Games distributed using the freemium business model are freely downloadable and playable. The main revenue for the games comes from virtual goods that can be purchased by players. Furthermore, many games include some form of advertisement (e.g. banners) that serve as a supplementary revenue stream.

In the freemium industry, similarly to other service industries such as telecommunications, the revenue that a player can generate is proportional to the duration of the relationship between the player and the game/service. 
Therefore, increasing player retention (i.e. the duration of the period before a player quits) is commonly considered an effective strategy for increasing lifetime value~\cite{Reinartz2003OnMarketing}.

This can be achieved in many ways, for example by producing more content for players in end-of-content situations or by adjusting problematic sections in the game that have shown to lead players to quit.
Another possible way, as shown by Milosevic\cite{Milosevic2015EarlyEleven}, is to identify the players that are likely about to stop to playing (i.e. churn) and target them with a personalised re-engagement initiative before they abandon the game.

This is challenging especially in non-contractual services such as freemium games.
For contractual services, such as telephone subscriptions or newsletters, the churn event is well defined, and corresponds to the moment when the contract expires or is cancelled. 
However, for non-contractual services, such as games or retail, there is not an explicit event that signals that a user stops using the service.

The most common way, as described by Hadiji et al.~\cite{Hadiji2014PredictingWild}, is to define the churn time as the time of the last event produced by a player before being inactive for a certain period of time.
The duration of the inactivity may be very different depending on the context: for example, if a player does not return to a freemium game after one week it is much more likely that he/she has churned compared to not returning to a clothing retail shop after a week. 
Formalising churn is therefore industry and time scale dependent and has to take into account the applicability to the business.

Regardless of the churn definition, churn prediction is currently actively researched in number of different industries including telecommunication providers~\cite{Prashanth2017HighIndustry, Hashmi2013CustomerClassification}, insurance companies~\cite{Zhang2017DeepService}, pharmaceutical companies~\cite{SIMION-CONSTANTINESCU2018DeepPrediction} and games~\cite{Lee2019GameData}.

Within games, a number of techniques have been employed for churn prediction ranging from a number of supervised learning models based on aggregated player data~\cite{Hadiji2014PredictingWild,Runge2014ChurnGames} to more recent works that try to leverage the dynamics for the player behaviour by using temporal data~\cite{Kim2017ChurnData}.

The main reason to use this kind of data is that the changes in the user behaviour leading up to the churn event are potentially more predictive than aggregated data.
Such an assumption is supported by a number of other recent studies on churn prediction in other industries~\cite{Grob2018ATime,Martinsson2017WTTE-RNNCase,Tamassia2017PredictingGame}.

However, since these temporal based methods focus on the dynamics of the player behaviour in a limited time window, they are unable to capture the baseline behavioural patterns of the players and assume that a specific sequence of events determines churn independently of the player's history and context.

Inspired by the work of Leontieva and Kuzovkin~\cite{Leontjeva2016CombiningClassification} on combining static and dynamic features for classification, in this article we investigate how both sequential and historic aggregated data about the player behaviour can be used in churn prediction models. The hypothesis behind this study is that static data about the player could serve as context to interpret the dynamics of the player behaviour.
For this reason, we evaluate a number of different architectures that can be used to combine the two types of data and we showcase the results in a comparative analysis based on data from a commercial free-to-play game.

The structure of the article is as follows: first the state-of-the-art methods for churn prediction will be presented in section \ref{sec: related work}. 
Next the churn definitions and methods that we will use in this article are introduced in \ref{sec: methods}. 
The evaluation parameters and data will be described in \ref{sec: evaluation}, and the results of the evaluation is shown in \ref{sec: results}.
Lastly, section \ref{sec: discussion} is a discussion of the results, followed by the conclusion in section \ref{sec: conclusion}.

\section{RELATED WORK}
\label{sec: related work}

While the concept of customer churn has been used in research for many years, the first examples of models for churn prediction start to be published in the late nineties and the early two thousands~\cite{Masand1999CHAMP:Prediction,Mozer2000ChurnIndustry}.
In their works, Masand et al. and Mozer et al. employ artificial neural networks (with slightly different topologies and feature selection methods) to predict whether a customer will cancel their telephone subscription or not. Other methods, such as decision trees \cite{Wei2002TurningApproach}, support vector machines (SVM) \cite{XIA2009ModelMachine} and logistic regressions, have also been used extensively for churn prediction \cite{Kamakura2006DefectionModels, Hashmi2013CustomerClassification, DahiyaKanikatalwar2015CustomerReview}, with many variations detailed in \cite{Santharam2018SurveyTechniques}.

All of the aforementioned methods for churn prediction attempt to assess the likelihood of a customer to churn based on their past behaviour expressed as a static summary of their state. These models assume that conditions leading to a churn event are based only on a given state of the customer rather than the way customer reached that given state.
This means that, for instance, two players with the same average number of hour played per day would be classified in the same way even if one of the two is playing increasingly more while the other is progressively stopping.

To capture this type of difference, the inputs to the model need to incorporate a temporal dimension. This dimension can be either approximated (e.g. incorporating trend and standard deviation to the aggregated measure) or the model can process the inputs as time series.
Castro and Tsuzuki~\cite{Castro2015ChurnApproach}, for instance, analyse a number of methods to approximate the dynamics of the customer behaviour using different forms of frequency-based representations.

If a feature can be arranged into time-sequential bins (e.g. hourly score, daily time played, monthly minutes on call), a more complete representation of the dynamic behaviour can be expressed in the form of a multi-variate time series, in which each sample of customer behaviour is described as a matrix with $n_{t}$ rows and $n_{f}$ columns, where $n_{t}$ is the number of time steps/length of time-series and $n_{f}$ is the number of features.

Prashanth et al.~\cite{Prashanth2017HighIndustry} present two different ways of processing time series using machine learning models. In their compararive study, in one of the case, they employ a long short-term memory (LSTM)~\cite{Hochreiter1997LongMemory} recurrent neural network using the data directly as time series. In the other case they flatten the multivariate time-series matrix into a single vector with length $n_{t} \cdot n_{f}$. By flattening the time-series, additional static features such as days since last usage and age can be appended to the vector. This vector is then used as input to non-sequential models such as a random forest classifier (RF) and a deep neural network.

A similar approach is used in \cite{Khan2017SequentialPerspective} where the static features (e.g. user age) are repeated for each month for the sequential models. While the performance of the different models is comparable, in both articles the RF outperformed the LSTM approach in terms of area under the curve (AUC). Another architecture that allows using sequential data is Hidden Markov Models (HMM) which is used in \cite{Rothenbuehler2015HiddenPrediction}.

One issue with framing churn prediction as a binary classification problem is that we do not know if/when a customer churns in the future. Because this information is hidden in the future the data is said to be right-censored. 
So, instead of framing the churn prediction as a binary classification problem, methods such as survival analysis attempt to estimate the time to the next event of interest, for instance the return of the customer or cancellation of subscription.

Survival analysis is extensively used in engineering and economics, and popular methods include Cox Proportional Hazards Model~\cite{Cox1975RegressionLife-Tables} and Weibull Time To Event model~\cite{Anderson2006AModel}.
 Both methods have been also applied to churn prediction alone and in combination with other classifiers~\cite{Ishwaran2008RandomForests,Perianez2016ChurnEnsembles,Martinsson2017WTTE-RNNCase,Grob2018ATime}.

\subsection{Churn prediction in games}

In both a general industry and games context, the two main approaches for churn prediction consider the churn prediction task as either a classification or survival analysis problem.

In \cite{Perianez2016ChurnEnsembles}, Perianez et al. interpret churn prediction as a survival analysis problem and focus on predicting churn for high-value players using a survival ensemble model.
One of the first examples of treating churn prediction as a classification problem in games is the 2014 article by Hadiji et al.~\cite{Hadiji2014PredictingWild}.

In this work, the authors describe two different forms of churn classification problems, in which the algorithm is either trained to detect whether the player is currently churned (\textbf{P1}) or whether the player will churn in a given future period of time (\textbf{P2}).
Furthermore, they compare a number of classifiers based on aggregated gameplay statistics on both tasks on datasets from five different games, showing decision trees to be the most promising classifier.

In the same year, Runge et al.~\cite{Runge2014ChurnGames} present an article investigating how to predict churn for high value players in casual social games.
In this article, high value player are defined as the top 10\% revenue-generating players, the churn definition is similar to the one labelled as P1 by Hadiji et al.~\cite{Hadiji2014PredictingWild}, and the period of inactivity used to determine churn is 14 days.

A set of classifiers similar to~\cite{Hadiji2014PredictingWild} -- with the addition of support vector machines -- is evaluated on the dataset from two commercial games.
For the feed-forward neural network and logistic regression models it was found that 14 days of data prior to the churn event leads to the highest AUC. 

Furthermore, to include a temporal component in the model, sequences of the daily number of logins are processed through a Hidden Markov Model. The output of the HMM is then used as an extra input feature. The authors, however, find the the inclusion of the temporal data using HMM degrades the results and hypothesise this might be due to data over-fitting.

A Hidden Markov Model is also used by Tamassia et al.~\cite{Tamassia2017PredictingGame} in comparison with other supervised learning classifiers based on aggregated data. The comparative study, conducted on data from the online game Destiny\footnote{https://www.destinythegame.com/d1}, shows an advantage in processing the player behaviour as temporal data. 

Kim et al~\cite{Kim2017ChurnData} also investigate the predictive power of sequential data by evaluating an LSTM Neural Network model in predicting churn for new players.
In this work, the input data to the LSTM corresponds to a single time series containing the player score recorded every 10 minutes over 5 days; churn is defined as having no activity for 10 days after the first 5 days of observation.

The results show that the LSTM model is able to outperform both a one-dimensional convolutional neural network on the same time series data and traditional learning models (RF, Gradient boosting, logistic regression) in terms of AUC. A similar result is achieved also by the LSTM based model by YOKOZUNADATA in the churn prediction competition article by Lee at al.~\cite{Lee2019GameData}.

Outside of the context of churn prediction in games, Leontjeva and Kuzovkin~\cite{Leontjeva2016CombiningClassification} show in their article that a hybrid LSTM network combining aggregated and time-series data is capable of better churn prediction than methods using only one of the two data types or classical ensemble methods.

These results combined with the aforementioned results by the YOKOZUNADATA LSTM based model suggest that there is potential for hybrid LSTM networks to leverage the combination of aggregated an time-series data.
For this reason, in this article we present a comparative study of multiple hybrid architectures of LSTM to evaluate the best possible solution in a realistic churn prediction problem.

\section{METHODS} \label{sec: methods}

In this study, we compare a number of different hybrid LSTM architectures that combine time-series data with aggregated data against commonly employed LSTM neural network and random forest algorithms. In this section, we describe all the architectures, the algorithms and the settings employed, while in the next section, we describe the evaluation procedure.
However, before describing the algorithms, it is first necessary to define what definition of churn will be used to label the data for the algorithms training and evaluation. This choice motivates what kind of data is relevant and can be used and that, in turn, will also determine what kind of architectures can be tested.

\subsection{Churn prediction definitions}

In freemium games the relationship between a player and the game is typically non-contractual in nature because the user can stop playing the game without any notice. 
In this situation there is not clear churn event, like a customer cancelling a subscription. 
For this reason, different research works have slightly different definition of churn; however, they all agree that a player can be considered churned if inactive for a long enough period of time~\cite{Lee2019GameData}.

\begin{figure}[t!]
    \centering
    \includegraphics[width=\columnwidth]{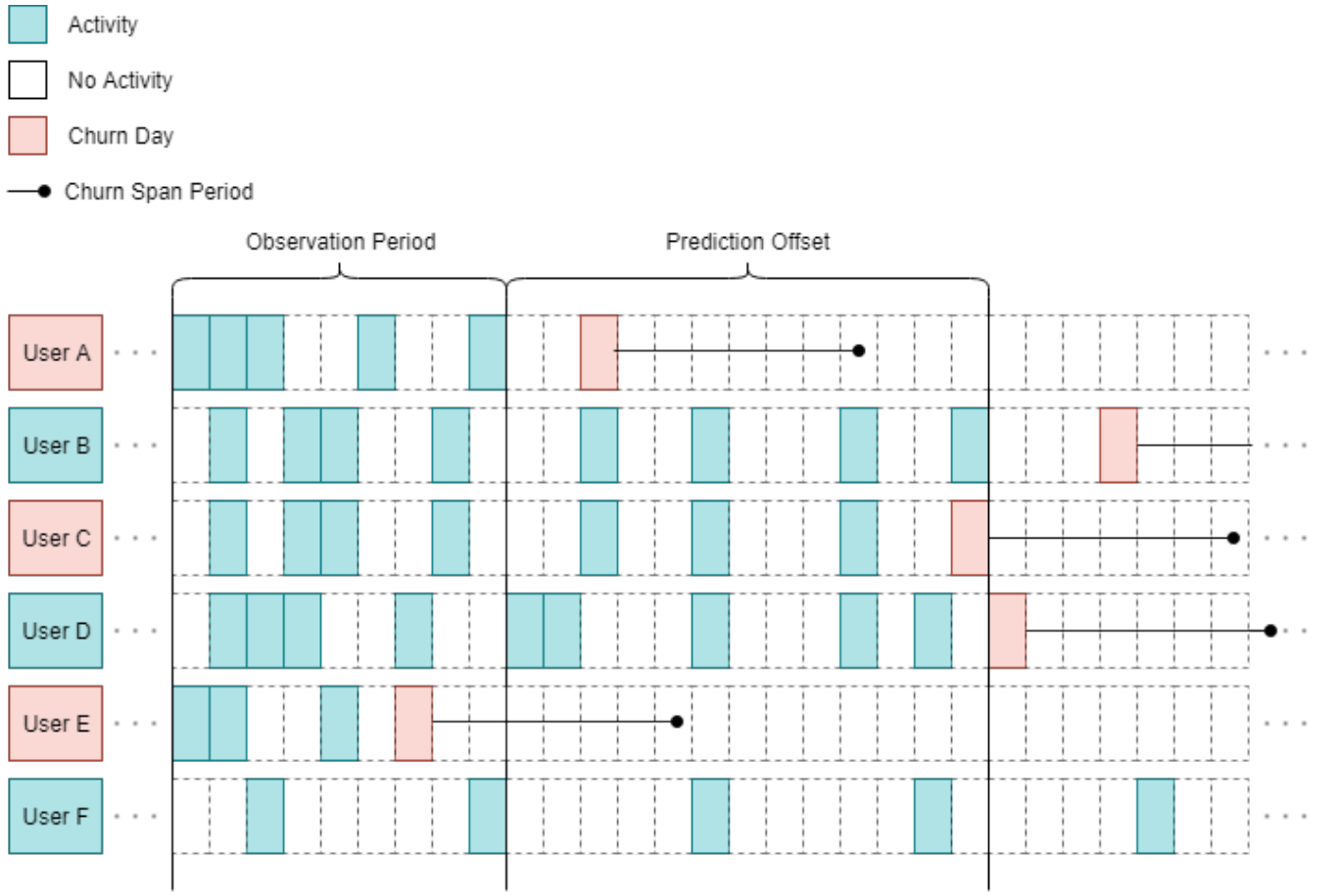}
    \caption{Depiction of the churn definition used to label the data. The predictions are made the day after the last day of the observation period/first day of the prediction offset. In this example user A, C and E are labelled as churners because their churn dates -- i.e. the last active day before a period of inactivity (churn span period) -- happen before the end of the prediction window. Even though user B and D have a churn date, they are labelled as non-churners because it happens after the prediction window. This is not a problem since their churn will be detected at a later prediction when it is appropriate to reengage them. User F is continuously active and does not churn either. Image courtesy of \cite{Heiberg-Iurgensen2019ChurnLSTM}.}
    \label{fig: churn definition}
\end{figure}

In this work, we define a churn event as the last event generated by a player before a period of inactivity. The churn prediction task, similar to the \textbf{P2} definition in \cite{Hadiji2014PredictingWild}, consists in predicting whether  churn event will occur in the next prediction period (e.g. the week following the prediction).
Figure \ref{fig: churn definition} show a number of examples of patterns of player activity and explains whether the players are considered churned or not according to our definition.

A second aspect of the churn classification task that we need to specify is which player is this model targeted at.
Kim et al.~\cite{Kim2017ChurnData} describe a model aimed at predicting churn for new players, while Runge et al.~\cite{Runge2014ChurnGames} and Perianez et al.~\cite{Perianez2016ChurnEnsembles} focus on high-value players. In contrast, the model we propose in this study is aimed at any player that is currently active. This means that at the time of prediction, the model can be applied to any player which has shown some activity (e.g. has performed at least one action) within the previous 14 days. 
This time window has been selected as 14 days is also the length of the input data time window. Which, in turn, has been chosen based on the periods selected in the literature. The period duration had to be a multiple of 7 days based on the periodicity of the players' behaviour in the game used for training and prediction.

A third aspect necessary to define is how long a player needs to be inactive before being labelled as churned. Choosing the duration of this period is a trade-off between finding actual churners versus players just taking a break.
Because of the aforementioned weekly periodicity, a minimum requirement for inactivity duration should be at least one week.
The maximum duration is not clear cut and can be chosen from a business perspective.
For example, if the cost of reengaging churning players is low, a short inactivity period can be chosen; however, at the same time a short inactivity period could lead the algorithm to label as churned a lot of player who would return later.
In this article the churn span period, i.e. duration of inactivity before being labelled as churner, is set to be 30 days.


 Lastly, in order to create actionable predictions, a sliding offset window from the prediction date (end of observation period) is used in which the churn can happen, similar to the \textbf{P2} definition in \cite{Hadiji2014PredictingWild}.
This allows for preemptive actions to be taken when a player about to churn, instead of when he/she has already churned. The length of the prediction offset window is 7 days.

\subsection{Models}

With the aim of finding the most effective way to combine time-series and aggregated player behaviour data, we include in the study three models which only use the sequential data are used as a baseline, and a number of different hybrid architectures. All implemented algorithms are based on either the Keras Deep Learning library\footnote{https://keras.io/} for Neural Networks or scikit-learn\footnote{https://scikit-learn.org/stable/} for the random forests and the evaluation heuristics.

\begin{figure}[t!]
    \centering
    \textbf{(A)} \textup{Baseline LSTM}\par\smallskip
    \includegraphics[width=\columnwidth]{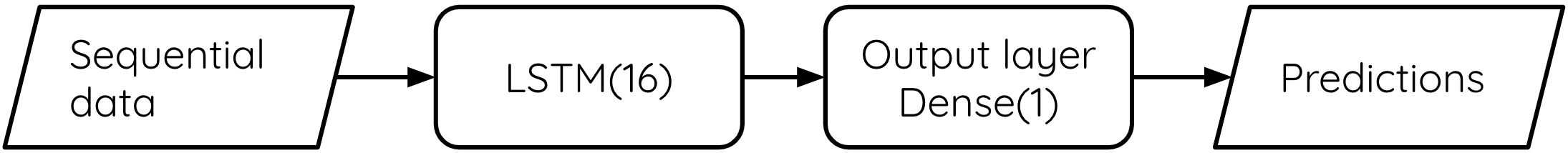}

\par\medskip\hrule
\par\medskip
    \textbf{(B)} \textup{LSTM + Aggregated/Prediction}\par\smallskip
    \includegraphics[width=\columnwidth]{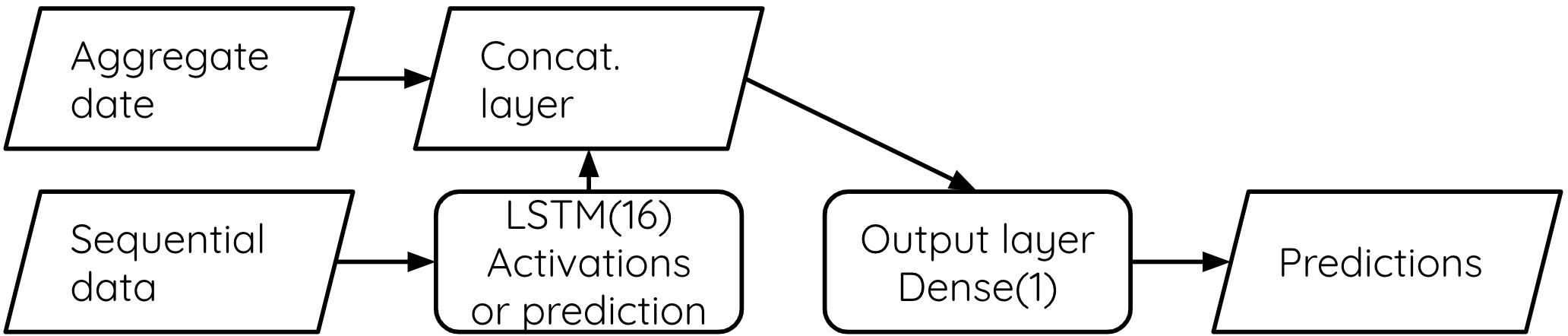}
\par\medskip
\hrule
\par\medskip

    \textbf{(C)} \textup{Hidden State LSTM}\par\smallskip
    \includegraphics[width=\columnwidth]{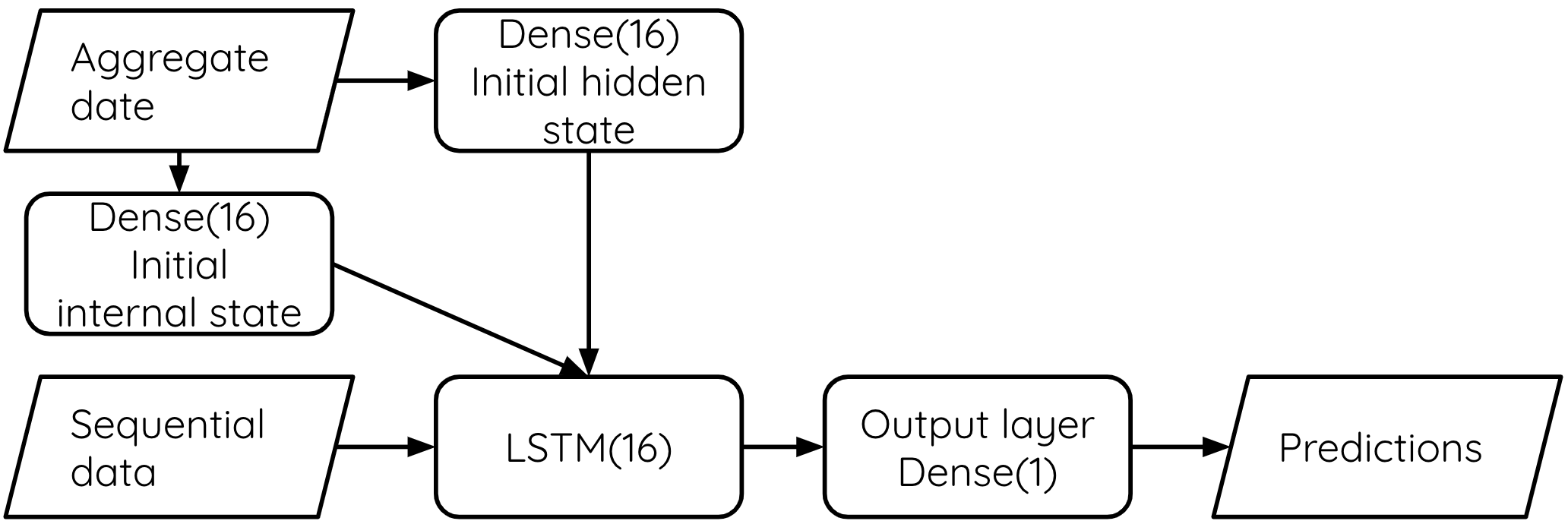}
    
    \caption{Structure of some of the different architectures tested in this article. \textbf{(A)} is a simple LSTM setup and serves as a baseline. \textbf{(B)} merges the aggregated data with either the activations of the LSTM or the churn prediction of the LSTM (baseline LSTM) which is then fed into the output layer. \textbf{(C)} connects the aggregated data with two separate dense layers which then serve as the initial hidden and internal cell states for the LSTM. The output from the LSTM cell is then fed into the output layer, similarly to the baseline LSTM model.}
    \label{fig: model architectures}
\end{figure}

The first two baseline models are a random forest classifier and a feed-forward neural network. Because these models cannot handle sequential data, the sequences are flattened into a single vector. 
The last baseline model is shown in Fig. \ref{fig: model architectures} (A). It consists of an LSTM layer to handle the sequential data with an output dimension of 16. 
Heuristically using a larger dimension did not improve the predictions and typically caused the model to overfit.
The LSTM layer uses the default settings of Keras -- i.e. the activation function is a hyperbolic tangent and the recurrent activation function is a hard sigmoid.

The hybrid models tested in this study include the architecture that is best performing in \cite{Leontjeva2016CombiningClassification}, which we label as \textit{LSTM + Aggregated}.
As shown in Fig. \ref{fig: model architectures} (B), this hybrid model relies on the temporal model to generate features from the time-series. 
The generated output is then concatenated with the static data and fed into a final classifier.

These LSTM output can either be the LSTM activations, the log-likelihood of belonging to each class or ratios of the likelihoods.
While all the hybrid models performed well in \cite{Leontjeva2016CombiningClassification}, the setup using LSTM activations generally performed better when many samples were used ($>5000$) and the sequence lengths were around 15 or longer; therefore, we use this configuration.
The final classifier uses a fully connected network output layer with 1 unit using a sigmoid activation function.

On top of this architecture, three other configurations are included in this study: \textit{LSTM Predict + Aggregated}, \textit{LSTM Hidden State} and \textit{Static in LSTM}.

The first one (\textit{LSTM Predict + Aggregated}) is a modified version of the \textit{LSTM + Aggregated} model that uses the LSTM prediction instead of the activation. The final sigmoid output of the LSTM serves as one of the inputs to the final classification layer together with the aggregated features. This architecture behaves similarly to the \textit{ensembles} described by Leontjeva and Kuzovkin~\cite{Leontjeva2016CombiningClassification} as the two classifiers operate independently.

In the \textit{LSTM Hidden State} model, the static input is used to set the initial states of the LSTM (see Fig. \ref{fig: model architectures}). 
This is done by feeding the input data into two separate dense layers with linear activations which correspond to the initial hidden state and initial internal cell state of the LSTM. 
Since the number of neurons in these layers must match the number of units in the LSTM, 16 units are used for the dense layers.

Finally, in the \textit{Static in LSTM} model, the static features are modelled as time-series with a constant value over time. These constant series, together with dynamic features, are used as inputs to an LSTM model as suggested by Khan et al.~\cite{Khan2017SequentialPerspective}
Otherwise, the structure of the LSTM network is the same as the baseline LSTM model. 

All the neural networks are trained using binary cross-entropy as a loss-function and an Adam optimiser. Early stopping is also utilised and uses the model weights from the best epoch if there are no improvement in the validation loss after 10 epochs.

\section{EVALUATION} \label{sec: evaluation}

\begin{figure}[t]
    \centering
    \includegraphics[width=0.45\columnwidth]{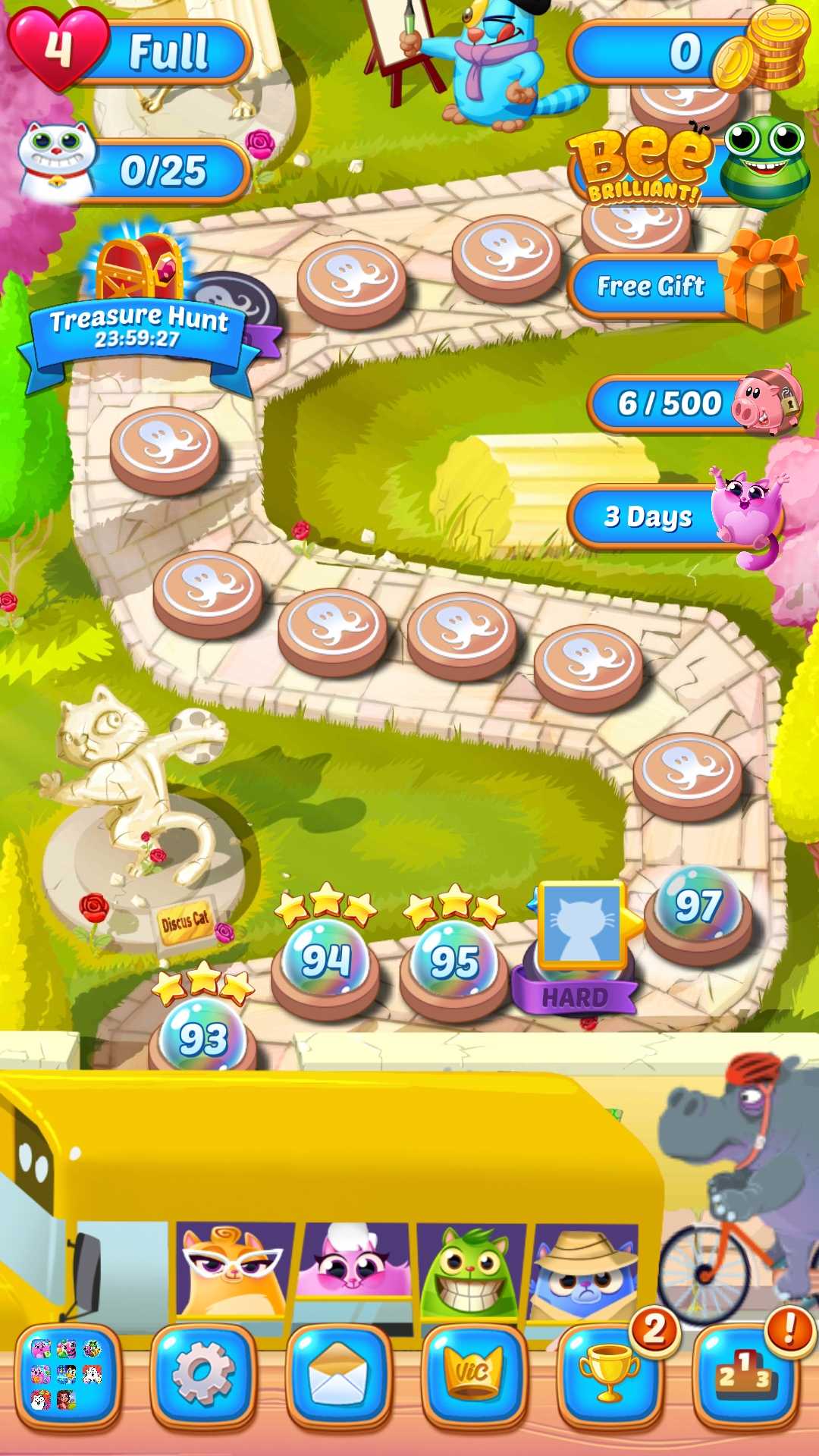}
    \includegraphics[width=0.45\columnwidth]{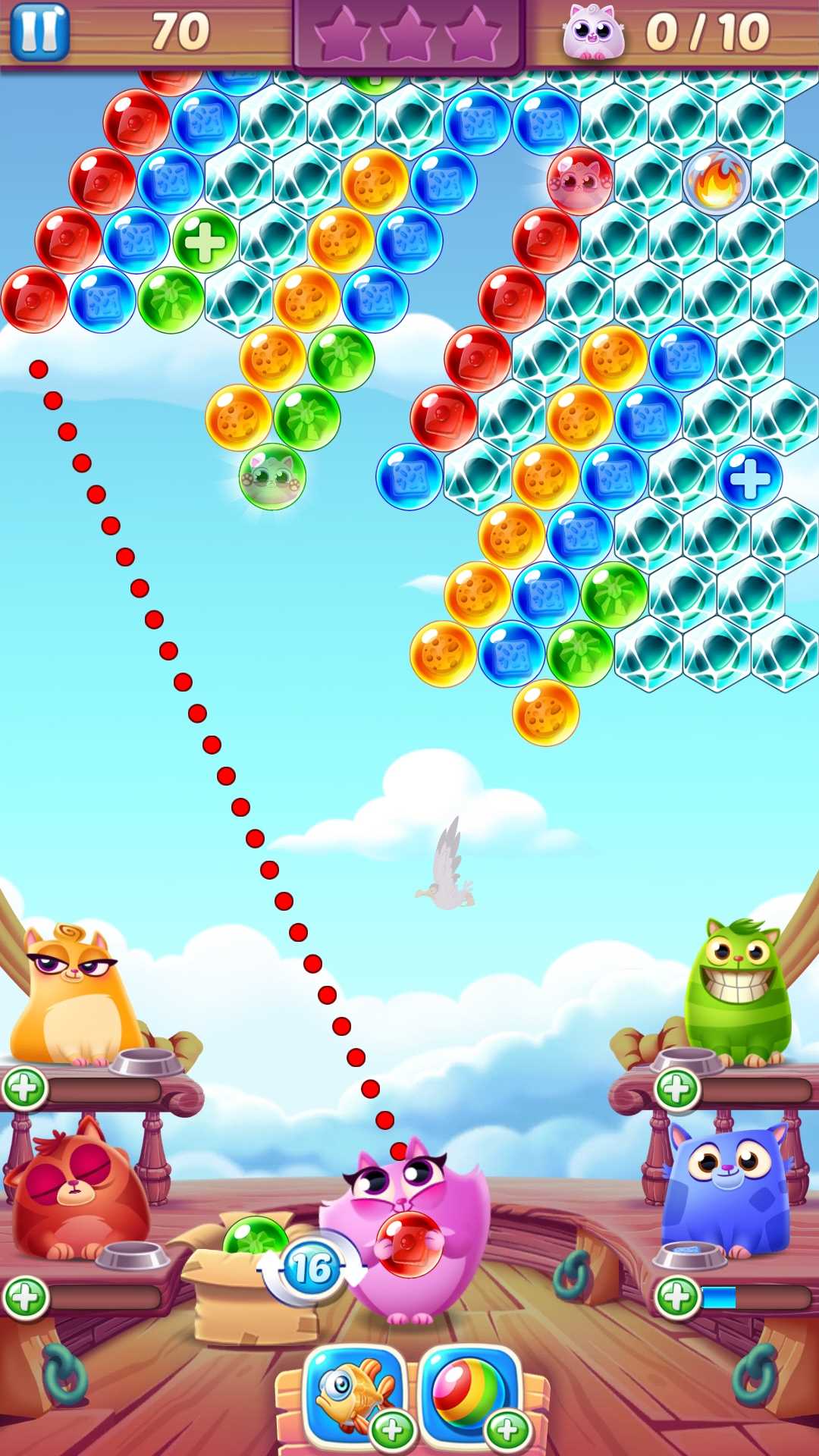}
    \caption{In-game screenshots from the mobile casual game Cookie Cats Pop, a pop shooter game for Android and IOS. The left image shows the world map where level progression can be seen (here it is level 96) and how in-game events appear (such as treasure hunt). The right shows an example of a level.}
    \label{fig: ccp screenshot}
\end{figure}

All the models described in the previous section are evaluated on the same churn prediction task; the data used for this test contains player logs from a casual mobile pop shooter game by Tactile Games called Cookie Cats Pop (Fig. \ref{fig: ccp screenshot}).
In these type of games the user typically has to complete levels over a linear or semi-linear progression and each level is composed by a different puzzle with the same core mechanics; in this case, the player has to shoot a  number of bubbles towards other bubbles to compose areas of the same colour and gather points. Various boosters, such as bonus actions or clearing the game board, can be used before or during the game as help to finish the level.

The initial dataset contains player behaviour data from 2018-08-01 to 2019-03-04. However, because we cannot know whether a player has churned until the inactivity period and prediction offset period have passed, the latest data is at least $30+7 = 37$ days before the upper-bound date.

The models are trained on two types of input data: \textit{aggregated data}, which summarises the characteristics of the player during the last 6 months up to the moment of prediction, and \textit{temporal data}, which contain time series describing daily summaries of the players' behaviour.

The temporal data contain both features that describe the activity level of the players and data that that reflect skill level of the player. The selection is based on the features included in \cite{Kim2017ChurnData} and \cite{Prashanth2017HighIndustry}.
In total, the following ten different features have been selected:
\begin{itemize}
    \item \textsc{activity}: 1 if player was active, otherwise 0
    \item \textsc{gameStarted}: number of times game/app was opened
    \item \textsc{missionStarted}: number of missions started
    \item \textsc{missionMovesUsed}: sum of moves used
    \item \textsc{pointsPerMission}: average points per mission
    \item \textsc{movesPerMission}: average moves used per mission
    \item \textsc{missionCompleted}: number of completed missions
    \item \textsc{missionCompletedFraction}: fraction of completed missions
    \item \textsc{missionFailed}: number of failed missions
    \item \textsc{converted}: 1 if in-app purchase, otherwise 0
\end{itemize}
Each training record is composed by these ten features, each feature has one daily entries for each of the previous 14 days (observation period).

The aggregated data contain features to describe the characteristics of the players and give context to the classifier to interpret the temporal data.
These features include game-specific metrics such as amount of in-game currency used, game feature/event participation and booster usage, but also aggregations of general playing patterns (e.g. number of active days, minutes played per day and max level reached).
The features can be grouped into the following categories:

\begin{itemize}
    \item \textit{Player description} which contain general descriptions of the player and consists of \textsc{fb-connected}, \textsc{monthssinceinstall}, \textsc{num-activedays} and \textsc{maxlvl}
    \item \textit{Player behaviour} which describes how the player behaves in-game and consists of \textsc{minutesplayed-sum}, \textsc{minutes-perday-avg}, \textsc{gamestarted-sum}, \textsc{levelstarted-sum}, \textsc{completionrate}, \textsc{abandonedrate}, \textsc{coinsused}, \textsc{coinused-perlevel}, \textsc{coinsreceived}, \textsc{continuesused-perlevel}, \textsc{boostersused-perlevel}, \textsc{transaction-sum}, \textsc{sum-spend}, \textsc{total-spend} and \textsc{progressionrate}
    \item \textit{Progression} which describes progress in different game modes and consists of \textsc{daily}, \textsc{main}, \textsc{onelife-challenge}, \textsc{social-challenge}, \textsc{tournament}, \textsc{treasurehunt}, \textsc{hot-streak}, \textsc{level-dash}, \textsc{levelrush} and \textsc{startournament}
    \item \textit{Platform} which describes what device player is using and consists of \textsc{android}, \textsc{fireos}, \textsc{ios} and \textsc{kindle}
    \item \textit{Acquisition channel} which describes how the player got invited to the game and consists of \textsc{acquired}, \textsc{crosspromoted} and \textsc{organic}
\end{itemize}
In total 22 features are used which expand to 36 features using one-hot encoding on categorical features.

As argued in the previous section, we use an observation period of 14 days, churn inactivity period of 30 days and a prediction offset window of 7 days.
Defining churn this way yields a data set with ~65\% non-churners and ~35\% churners. 
While methods such as  over- or under-sampling or bootstrapping can be used to deal with class imbalances, ensuring an even class distribution does not guarantee a better result, especially in a churn setting and when using AUC as the evaluation metric \cite{Burez2009HandlingPrediction}.
No further action is therefore taken to deal with the class imbalance.

\begin{figure}[t]
    \centering
    \includegraphics[width=1\columnwidth]{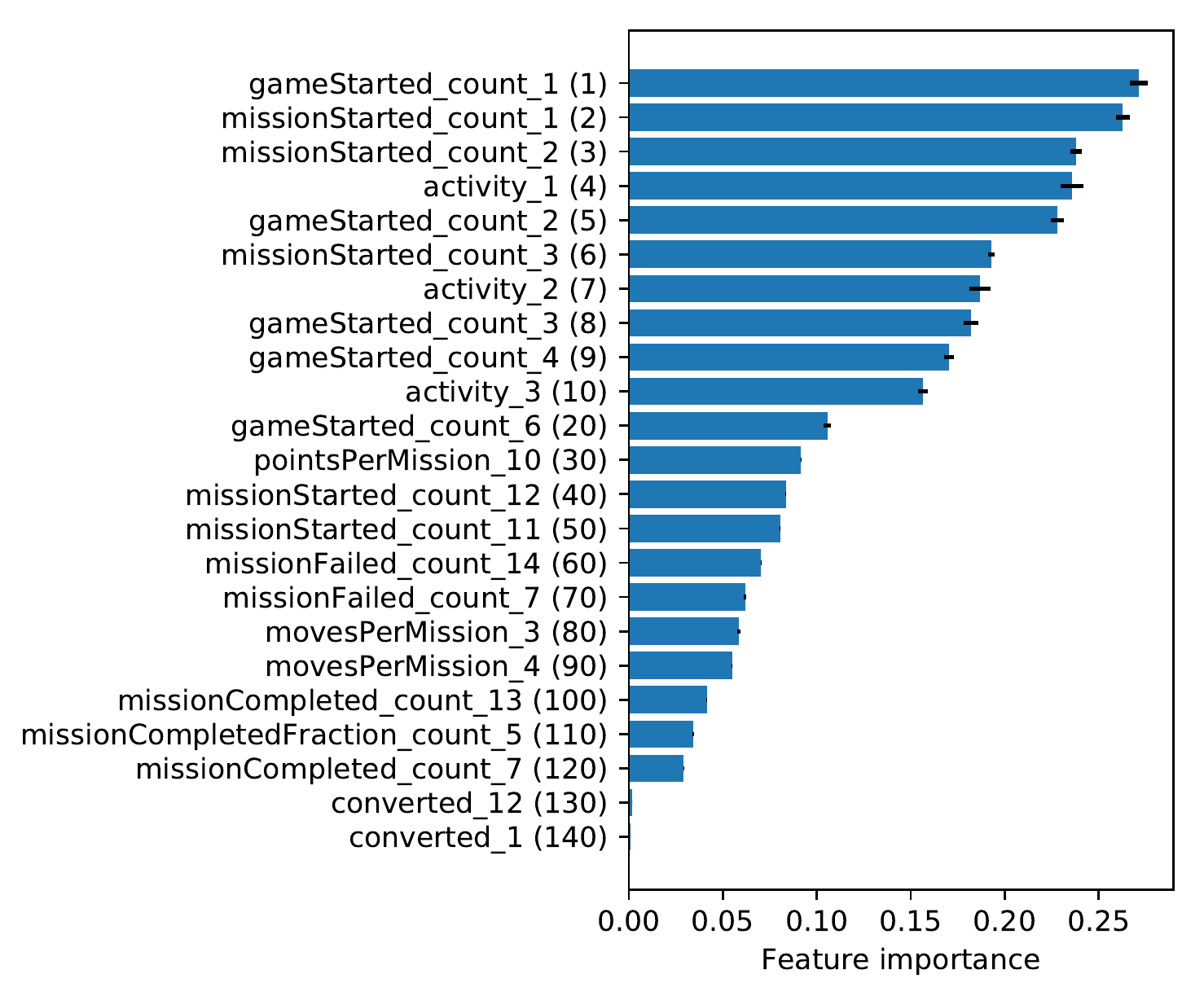}
    \caption{Feature importance of the baseline random forest model. Note that not all 140 features are shown, only the 10 most important features followed by every tenth feature. The number in the parenthesis indicates the order of importance. The suffixed number indicates number of days ago, where 1 is the most recent date.}
    \label{fig: feature importance}
\end{figure}

In order to gather a diverse data set covering a long enough period of time, 8 sampling dates that are each 18 days apart are chosen.
This ensures data for every week day is included and that the observation periods do not overlap. A player may be included in multiple sampling dates, but since there is no overlap in the observations it is assumed that the behaviour is independent. 
Each date has approximately 250,000 records resulting in a total data set of 2,284,238 records with 814,822 unique players.

For the evaluation of the different approaches, a 10-fold cross validation is performed.  In each fold three evaluation parameters are used: the area-under-curve score (AUC) of the receiver operating characteristic curve (ROC), the accuracy and the F1 score.

The ROC curve is a graph of the true positive ratio over the false positive ratio at different classification thresholds. The higher area-under-curve, the fewer false positives to true positives.
An AUC score of 1 is therefore the highest possible score while 0.5 corresponds to a random guessing model. 
Although using AUC with imbalanced data sets may not give a complete idea about the model performance \cite{Bekkar2013EvaluationSets}, the method is independent of choice of classification threshold and thus useful for a non-biased comparison between models.

Accuracy is a measure of how many correct predictions out of all the samples, i.e. $\textup{acc} = \frac{TP + TN}{N}$, where $TP$ is the number of true positives, $TN$ is the number of true negatives and $N$ is the number of samples. 
While accuracy is not a good metric on very imbalanced data sets, it is often used in literature and thus included for comparison.

The F1 score is the harmonic mean of the precision and recall of the model. Precision refers to the ratio of true positives to classified positives, and recall is a ratio that describes the number of true positives to actual number of positives.
Since the F1 score gives equal weight to precision and recall it can be used to measure the all around performance of the model.

Accuracy and F1 score require a binary classification so a classification threshold of $>0.5$ is used to label a player as churning.

\subsection{Results} \label{sec: results}

\begin{table}[t]
\caption{Model results. Number in parenthesis is the two sigma uncertainty on last significant digits. The models with the best performance are highlighted in bold.}
\label{table: model results}
\begin{center}
\begin{tabular}{l|ccc}
\textbf{Model} & \textbf{AUC} & \textbf{F1 score} & \textbf{Accuracy} \\[0.5ex]
\hline \\[-2ex]
Baseline RF & \multicolumn{1}{l}{0.8405 (21)} & \multicolumn{1}{l}{0.6414 (33)} & \multicolumn{1}{l}{0.7749 (21)} \\
Baseline ANN & \multicolumn{1}{l}{0.8559 (18)} & \multicolumn{1}{l}{0.6771 (54)} & \multicolumn{1}{l}{0.7868 (19)} \\
Baseline LSTM & \multicolumn{1}{l}{0.8592 (18)} & \multicolumn{1}{l}{0.6795 (48)} & \multicolumn{1}{l}{0.7900 (16)} \\
\textbf{LSTM + Aggregated} & \multicolumn{1}{l}{0.8729 (19)} & \multicolumn{1}{l}{0.6929 (50)} & \multicolumn{1}{l}{0.8013 (16)} \\
\textbf{LSTM Predict + Aggr.} & \multicolumn{1}{l}{0.8711 (20)} & \multicolumn{1}{l}{0.6898 (60)} & \multicolumn{1}{l}{0.8000 (17)} \\
\textbf{LSTM Hidden State} & \multicolumn{1}{l}{0.8741 (20)} & \multicolumn{1}{l}{0.6953 (30)} & \multicolumn{1}{l}{0.8023 (20)} \\
\textbf{Aggregated in LSTM} & \multicolumn{1}{l}{0.8737 (22)} & \multicolumn{1}{l}{0.6927 (34)} & \multicolumn{1}{l}{0.8020 (19)} \\
\end{tabular}
\end{center}
\end{table}

The results of the evaluation are shown in Table \ref{table: model results}.
Of the baseline models, the LSTM is better in terms of all three metrics, with significant differences that are larger than the uncertainties. The NN and RF classifier had very similar performances. 
Since the data is sequential in nature it is perhaps not surprising that models that are designed to deal with such data also performs better. However, it is a good test of the validity of using sequential data for churn.

In order to extract some information about the behaviour leading to churn, the RF model can be used to extract the feature importance, which is shown in Fig. \ref{fig: feature importance}. It can be seen that the three most important features are the most recent values for number of missions started, number of times game has been opened and whether a player was active -- all values which reflect play time. The least important features were all whether the player had converted.
These results are similar with what was found in \cite{Kim2017ChurnData} for other mobile casual games and in \cite{Prashanth2017HighIndustry} for telecom data.

\begin{figure}[h]
    \centering
    \includegraphics[width=1\columnwidth]{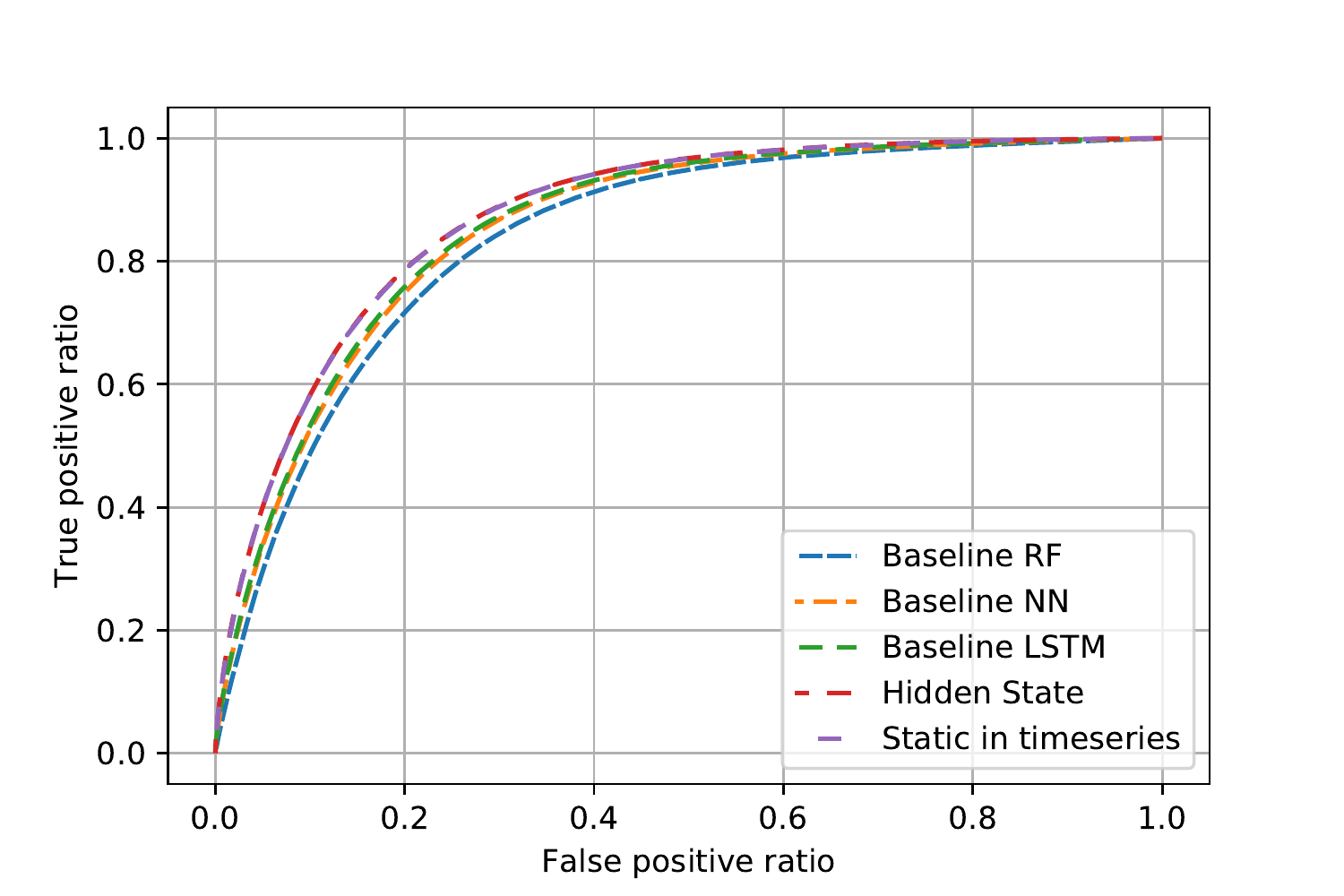}
    \caption{ROC curves for selected models.}
    \label{fig: roc curve}
\end{figure}

The LSTM Hidden State model and the one using static features in the sequential data have the best performance on all the parameters of all the tested architectures. However, the confidence intervals of the evaluation metrics of the different models overlap. It is worthwhile to note, though, that the training time of the Hidden State model is about one third faster than the static one, while the LSTM + Aggregated model was only slightly faster than the Hidden State model. In a business setting, the baseline LSTM model may be the fastest to both train and implement but it is only a small increase in complexity to use the Hidden State or LSTM + Aggregated model, which may then allow for use of domain-specific features that can boost the model performance for different user segments (e.g. level reached in games or in-game event participation). The best strategy may therefore be to use either the Hidden State or LSTM + Aggregated model and then tune the hyper-parameters.

A reason why the Hidden State and Aggregated in LSTM models appear to perform better than the LSTM + Aggregated models may be because the static features are not directly used in LSTM part of the latter models. It therefore limits what kind of features the LSTM can extract resulting in slightly worse performance. 
The ROC curves of some of the models are shown in Fig. \ref{fig: roc curve}, where it can be seen that the Hidden State and Aggregated in LSTM are better than the baselines but otherwise very similar.

We have also briefly tested the models only on converted users. These samples make up 10\% of the data set, and about 20\% of have a positive churn label (this number is 35\% for the overall data set). While the AUC and accuracy generally increased slightly, the F1 score decreased.

\section{DISCUSSION} \label{sec: discussion}

The results show that including the aggregated data increases the performance compared to the baseline LSTM and the improvement is comparable to using an LSTM over an RF or NN.
Interestingly, including static features in the time series (\textit{Static in LSTM}), and thereby increasing the number of sequential features, did not decrease the performance.
This is only somewhat in agreement with \cite{Kim2017ChurnData} where including more than 4 features lead to either no or a small decrease in performance.
However, it should be noted that the method employed by Kim et al.~\cite{Kim2017ChurnData} is a gradient boosting method whereas in this this article we use an LSTM.

Small variations of each model were tested, including inputting the aggregated features into a dense layer first (like for the hidden state model), adding a dense layer just before the output layer and using more cell units. However, no significant differences were found.

While combining aggregated and temporal data shows and improvement in churn prediction accuracy, it is worthwhile to consider that the aggregated features used in this study include many game-specific parameters, such as in-game currency spent, participation in game-specific events and so on. Further investigations would be needed to assess the generalisability of the results in other games. For instance, by replicating the experiment on the datasets used in~\cite{Lee2019GameData}.

Additionally, being able to use the same architecture for other games with similar mechanics may be of particular interest to some companies since that will allow them to target even more players with relative ease.
A more general approach may therefore be to cluster the users based on domain specific heuristics and use this group information as aggregated inputs.
Different rates of activity can also be used, keeping it as general yet informative as possible.

\begin{figure}[t]
    \centering
    \includegraphics[width=1\columnwidth]{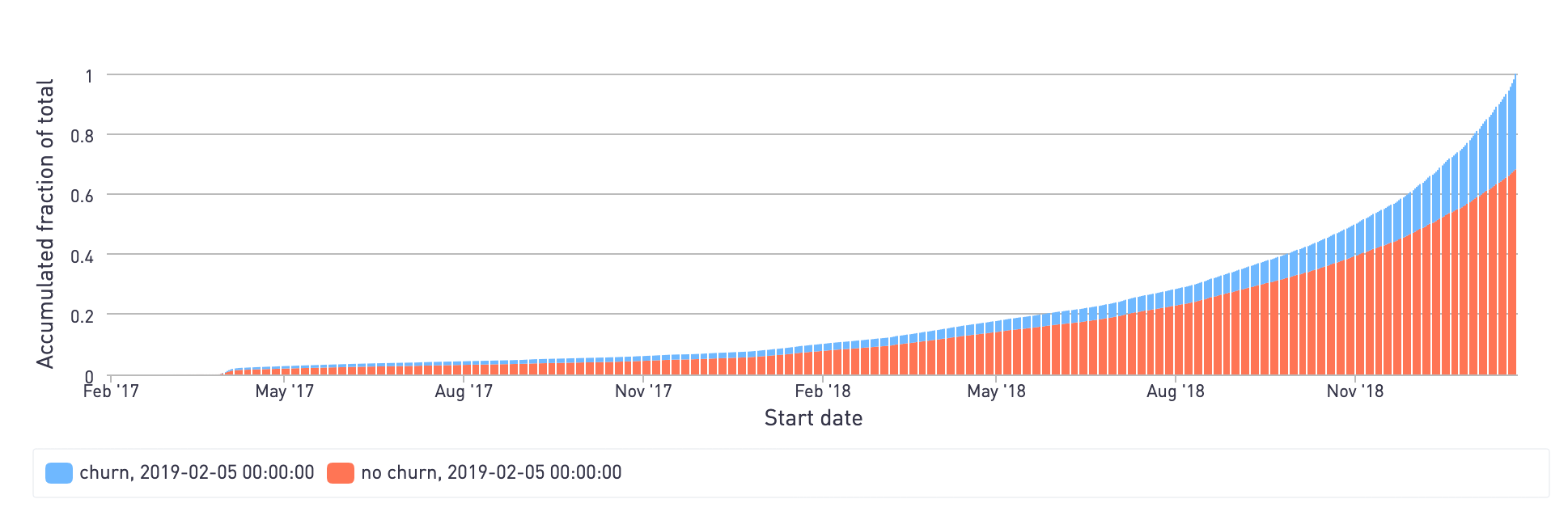}
    \includegraphics[width=1\columnwidth]{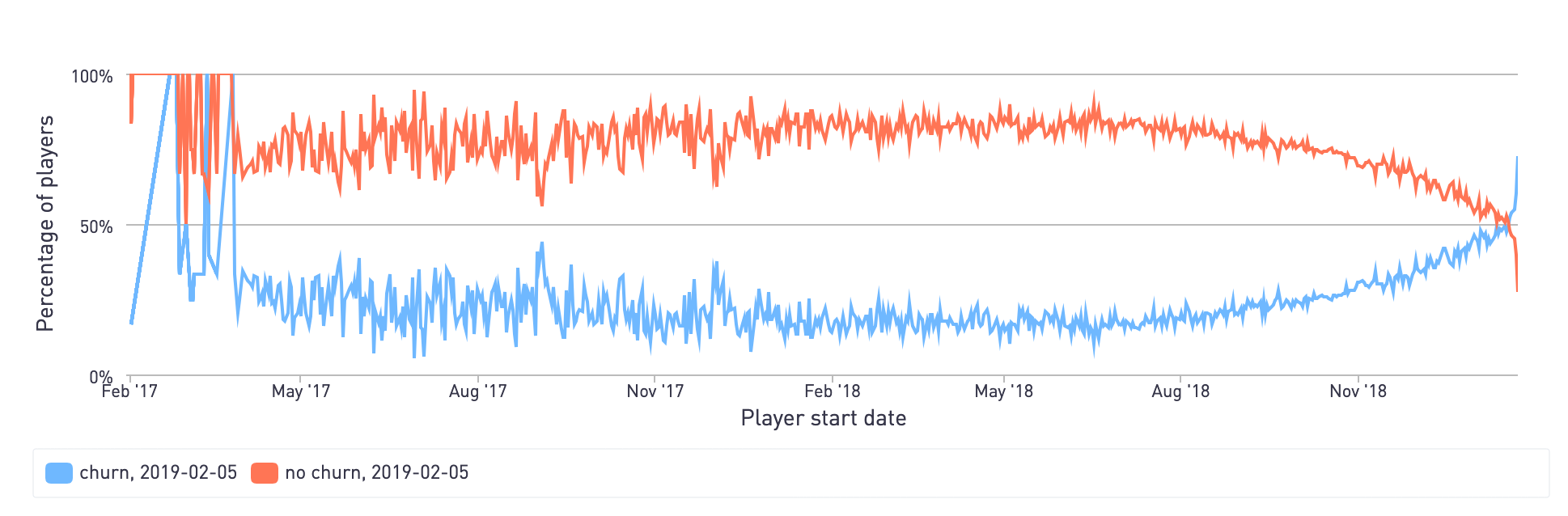}
    \caption{\textbf{Top:} number of player start dates split by churn label. \textbf{Bottom:} fraction of player start dates split by churn label. It can be seen that old players are generally less likely to be predicted to churn. The predictions were made on 2019-02-05 using the baseline LSTM model. A classification threshold of $>0.6$ is used, which is chosen from a rough estimate of the gain and probability of reengaging a true churner vs the cost of reengaging a falsely predicted churner.}
    \label{fig: player start dates}
\end{figure}

The small difference in model performance across architectures may suggest that activity features are enough to capture a churn signal in simple mobile casual games.

One thing that also affects the performance is how the data sampling is done.
The proposed method of using aggregated historical data and multivariate time series of the behaviour leading up to churn is a kind of supervised hierarchical temporal memory model.
This means that we choose the timescale (lifetime values and most recent 14 days) and binning window (e.g. daily aggregations) ourselves instead of in an automated way.
However, this way we may not capture all the temporal dynamics because we have explicitly chosen which dynamics to consider.
Indeed, looking at the player start dates (Fig. \ref{fig: player start dates}) split by churn label, it can be seen that of players starting roughly three weeks prior to the prediction date, a majority of them are churners.
Although it is still in the interest of the business to catch any player, the model may become specialised in predicting churn for new players and not learn to properly model the more profitable long-term players. 
This is also in line with the previous results: although the AUC was higher for converting players, this does not take into account the lower amount of identified churners, as reflected by the lower F1 score.

Some methods, such as the wavelet approach in \cite{Castro2015ChurnApproach}, allow for a bit more of an unsupervised approach in terms of data sampling, but the results are largely the same. 
In theory the input to a LSTM model can also be trained using complete life time sequences.
Additionally, stacked LSTMs may allow a model to learn different temporal dependencies \cite{PankajMalhotraLovekeshVigGautamShroff2015LongSeries}, which was also tested in this article but showed no improvement.

\section{CONCLUSIONS} \label{sec: conclusion}

In this article we presented and tested four neural network architectures for churn prediction that allow to combine aggregated historic data with sequential data.

The results show that combining the static features with either the LSTM prediction or activations showed an improvement over the baseline.
However, the best models were the ones that included the aggregated data in as some form of input to the LSTM -- either by setting the initial state or simply adding the static data to the time series.

As found in other articles, features that described the most recent activity carried the most importance for predicting churn. Using general activity patterns therefore form a good baseline and can be used across games.
However, some features used in the evaluation also included very game specific details. For this reason, we believe it is worth investigating some form of generalisation of the aggregated data, and we plan on investigating how to employ player profiling as an input in our future works on this topic.

Furthermore, a more dynamic data extraction scheme that can capture different temporal dynamics depending on player type may give an even better performance. Individual models for player archetypes will also give a better understanding of the predictions allowing for differentiated re-engagement strategies which will keep players, both old and new, passionate or absent, interested the game.

Lastly, for our future tests, we plan on expanding our suite of datasets by including all available data used in other articles such as~\cite{Lee2019GameData}.

\bibliographystyle{plain}
\bibliography{references}

\end{document}